# Machine Learning Methods for Gene Regulatory Network Inference

*Akshata Hegde[1,2], Tom Nguyen[1,2], Jianlin Cheng[1,2]\**

[1]*Department of Electrical Engineering and Computer Science, NextGen Precision Health, University of Missouri, Columbia, 65201, Missouri, USA*

[2]*Roy Blunt Nextgen Precision Health, University of Missouri, Columbia, 65201, Missouri, USA*

\*Corresponding author: chengji@missouri.edu

**Abstract**

Gene Regulatory Networks (GRNs) are intricate biological systems that control gene expression and regulation in response to environmental and developmental cues. Advances in computational biology, coupled with high-throughput sequencing technologies, have significantly improved the accuracy of GRN inference and modeling. Modern approaches increasingly leverage artificial intelligence (AI), particularly machine learning techniques—including supervised, unsupervised, semi-supervised, and contrastive learning—to analyze large-scale omics data and uncover regulatory gene interactions. To support both the application of GRN inference in studying gene regulation and the development of novel machine learning methods, we present a comprehensive review of machine learning-based GRN inference methodologies, along with the datasets and evaluation metrics commonly used. Special emphasis is placed on the emerging role of cutting-edge deep learning techniques in enhancing inference performance. The potential future directions for improving GRN inference are also discussed.

## 1. Introduction

Gene expression is the process by which genetic information synthesizes functional products, such as RNA and proteins, and is critical in all living organisms[1]. Proper regulation of gene expression is essential to ensure that genes are activated only when necessary and that their activity is properly controlled [3]. The regulation of gene expression is achieved through understanding the intricate interactions between genes and other molecules. In this effort, Gene Regulatory Networks have emerged as a strong tool[2].

Gene regulatory networks (GRNs) are complex systems that determine the development, differentiation, and function of cells and organisms, as well as their response to environmental stimuli [4][5]. GRNs consist of genes, transcription factors (TFs), microRNAs, and other regulatory molecules that interact with each other to control gene expression [6]. The regulatory interactions between these molecules can form complex networks that exhibit emergent properties, such as robustness and adaptability [7]. In its simplest form, a GRN is a network of genes and their regulatory interactions, which govern the expression of these genes in response to various cellular cues. It is worth noting that in this definition, a transcription factor (TF) is considered a special kind of gene that may regulate the expression of other non-TF or TF genes. Each gene in the network acts as a node, and the regulatory interactions between genes are represented by directed edges connecting these nodes[8].

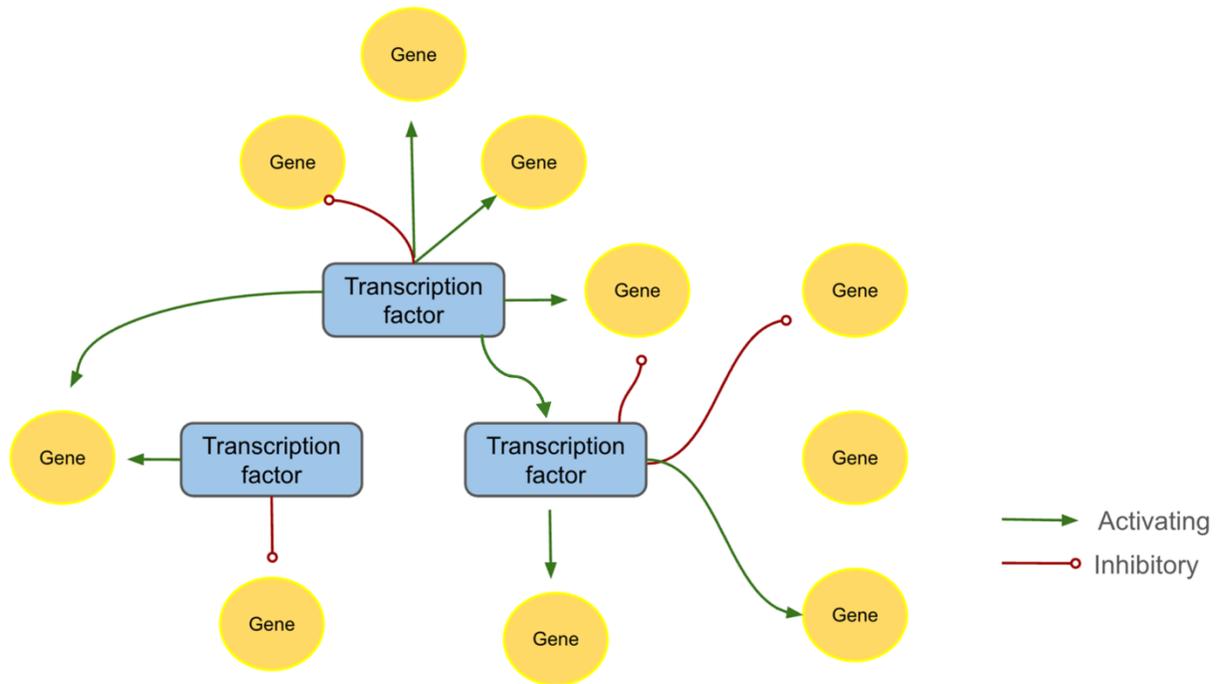

**Figure 1**. A simple example of GRN.

**Figure 1** shows a simple example of GRN [9]. The interactions (edges) in a GRN can be both activating and inhibitory, resulting in complex regulatory circuits that determine the expression of genes across different cellular states and in response to various environmental stimuli. The topology of a GRN can be influenced by a variety of factors, including gene duplication, mutation, and selection, leading to the evolution of novel regulatory mechanisms and gene functions[4]. Computational and experimental approaches can be used to study GRNs, providing insights into fundamental biological processes such as cell signaling, gene regulation, and protein-protein interactions [4][6]. Research in GRNs has significant implications for many fields such as systems biology, developmental biology, cancer biology, evolutionary biology, and personalized medicine[4].

Gene regulatory network inference or modeling is the process of identifying the interactions among genes that contribute to the regulation of gene expression. Over time, the study of GRNs has evolved from the early days of molecular biology to the current era of computational biology due to the generation and accumulation of huge amounts of multi-omics (e.g., genomics and transcriptomics) data that can be used to infer underlying gene regulation mechanisms.

The study of GRNs has a rich history that dates to the early days of molecular biology, when researchers first began to uncover the basic principles of gene regulation, such as the role of transcription factors in controlling gene expression[10]. In the late 1980s and 1990s, techniques such as DNA footprinting[11] and electrophoretic mobility shift assays (EMSAs)[13] were developed to identify transcription factor binding sites in DNA sequences[12][14]. The advent of microarray technology in the early 2000s allowed for large-scale studies of gene expression patterns [15], which paved the way for more advanced GRN modeling techniques.

In recent years, advancements in next-generation sequencing have dramatically increased the availability of high-throughput multi-omics data, boosting the development of gene regulatory network (GRN) inference methods. RNA sequencing (RNA-seq), for example, generates high-resolution gene expression data, offering a more detailed view than traditional microarrays [16][17]. The emergence of single-cell sequencing now allows researchers to profile gene expression and chromatin accessibility at the single-cell level, shedding light on cellular heterogeneity and developmental pathways [18]. Additionally, improved epigenetic profiling tools—such as chromatin immunoprecipitation sequencing (ChIP-seq) and assay for transposase-accessible chromatin sequencing (ATAC-seq)—have provided high-quality data for GRN inference. ChIP-seq identifies transcription factor binding sites [19], while ATAC-seq reveals chromatin accessibility patterns, both of which are crucial for modeling gene regulation [20].These breakthroughs in high-throughput data have expanded the scope of GRN inference methods, enabling the construction of more comprehensive and accurate models of gene regulation [21][22]. However, reliably inferring GRNs—i.e., uncovering the underlying biological regulatory mechanisms—from vast multi-omics datasets requires the development of sophisticated computational methods, particularly those based on artificial intelligence, along with rigorous benchmarking practices.

The DREAM (Dialogue on Reverse Engineering Assessment and Methods) projects were established to encourage the development of algorithms focused on GRN inference. These initiatives provide a suite of benchmark networks—derived from simulated *E. coli* and *S. cerevisiae* gene expression data—as well as real experimental datasets, facilitating large-scale, community-driven evaluations of algorithmic performance [33–35]. As a result of these efforts, GRN modeling has increasingly emphasized machine learning approaches that aim to predict regulatory network structures and gene expression dynamics with greater precision. These methods leverage high-throughput data to identify complex molecular interactions that drive gene regulation, surpassing earlier clustering-based approaches, which were limited in capturing significant transcriptional interactions [23].

Several classic machine learning algorithms have been widely applied in GRN inference, including Bayesian Networks, Structural Equation Modeling, Random Forests, Support Vector Machines, Gradient Boosting, Logistic Regression, and neural networks. More recently, deep learning has emerged as a transformative approach, offering powerful tools for modeling the complex, nonlinear relationships that characterize gene regulation. While multiple reviews have discussed GRN inference methods—some focusing on early computational approaches [24], others limited to transcriptomics-based methods [25], and a few addressing recent advances using chromosome structural data [26]—a comprehensive and up-to-date synthesis that integrates recent deep learning techniques across multiple data modalities remains lacking.

In this review, we aim to fill that gap by systematically categorizing state-of-the-art machine learning approaches for GRN inference, with a particular emphasis on the latest deep learning models. Unlike previous reviews, we not only classify methods based on algorithmic approaches, but also consider the types of data sources (e.g., transcriptomics, epigenomics, chromatin structure) and the specific forms of GRN inference they enable. This multidimensional framework is intended to provide researchers with a clearer understanding of current trends, emerging challenges, and future opportunities in the field.

**2. Machine Learning Methods for GRN Inference**

We categorize GRN inference methods broadly based on the type of machine learning methods i.e. supervised learning, unsupervised learning, semi-supervised learning, and contrastive learning methods.

**Table 1**. The categorization of 23 recent or representative machine learning methods for GRN inference.

| Algorithm Name | Learning Type | Deep Learning | Input Type | Year | Key Technology | Link |
| --- | --- | --- | --- | --- | --- | --- |
| GENIE3 | Supervised | No | bulk | 2010 | Random Forest | https://github.com/vahuynh/GENIE3 |
| SIRENE | Supervised | No | bulk | 2009 | SVM | http://cbio.ensmp.fr/sirene |
| GRADIS | Supervised | No | single cell | 2023 | Support Vector Machine | https://github.com/MonaRazaghi/GRADIS |
| DeepIMAGER | Supervised | Yes | single cell | 2024 | CNN | https://github.com/shaoqiangzhang/DeepIMAGER |
| DeepSEM | Supervised | Yes | single cell | 2023 | Deep Structural Equation | https://github.com/HantaoShu/DeepSEM |
| STGRNs | Supervised | Yes | single cell | 2023 | Transformer | https://github.com/zhanglab-wbgcas/STGRNS |
| RSNET | Supervised | Yes | single cell | 2022 | Graph Convolutional Net | https://github.com/zhanglab-wbgcas/rsnet |
| dynGENIE3 | Supervised | No | single cell | 2018 | Random Forest Modeling | http://www.montefiore.ulg.ac.be/ huynh-thu/dynGENIE3.htm |
| GRNFormer | Supervised | Yes | single cell | 2025 | Graph Transformer | https://github.com/BioinfoMachineLearning/GRNformer.git |
| AnomalGRN | Supervised | Yes | single cell | 2025 | Graph Anomaly Detection | https://github.com/ZZCrazy00/AnomalGRN |
| LASSO | Unsupervised | No | bulk | 2016 | Regression | https://github.com/omranian/inference-of-GRN-using-Fused-LASSO |
| ARACNE | Unsupervised | No | bulk | 2006 | Information Theory | https://califano.c2b2.columbia.edu/aracne |
| MRNET | Unsupervised | No | bulk | 2007 | Min. Redundancy/Info Theory | https://bioconductor.org/packages/release/bioc/html/minet.html |
| BiGRN | Unsupervised | Yes | bulk | 2022 | Bidirectional RNN | https://gitee.com/DHUDBLab/bi-rgrn |
| CLR | Unsupervised | No | bulk | 2007 | Mutual Information | https://bioconductor.org/packages/release/bioc/html/minet.html |
| GENECI | Unsupervised | No | bulk | 2023 | Evolutionary ML | https://github.com/AdrianSeguraOrtiz/GENECI |
| CVGAE | Unsupervised | Yes | single cell | 2024 | Graph Neural Network | None |
| GRN-VAE | Unsupervised | Yes | single cell | 2020 | Variational Autoencoder | https://bcb.cs.tufts.edu/GRN-VAE |
| BiRGRN | Unsupervised | Yes | single cell | 2022 | Bidirectional RNN | https://gitee.com/DHUDBLab/bi-rgrn |
| DeepMAPS | Unsupervised | Yes | scATAC/Multi-omic | 2023 | Heterogeneous Graph Transformer | https://github.com/OSU-BMBL/deepmaps |
| GRGNN | Semi-Supervised | Yes | single cell | 2020 | Graph Neural Network | https://github.com/juexinwang/GRGN |
| GCLink | Contrastive | Yes | single cell | 2025 | Graph Contrastive Link Prediction | https://github.com/Yoyiming/GCLink |
| DeepMCL | Contrastive | Yes | single cell | 2023 | CNN | https://github.com/lzesyr/DeepMCL |

**Table 1**. provides an overview of a list of various GRN inference algorithms categorized by their learning paradigms (supervised, unsupervised, semi-supervised, and contrastive learning), utilization of deep learning techniques, compatibility with bulk cell data, year of publication, and the core computational technologies employed. The list includes 14 recent deep learning methods as well as 9 other typical non-deep learning machine learning methods for GRN inference to be reviewed in this work. We focus on reviewing the recent representative deep learning methods, while considering some non-deep learning methods to provide a broad perspective of the field. Moreover, many methods that extend some popular methods in **Table 1** will also be discussed.

The diversity of approaches highlights the evolution of GRN modeling from classical machine learning methods (e.g., Random Forests, SVMs) to more recent deep learning frameworks including convolutional neural networks (CNNs), variational autoencoders (VAEs), graph neural networks (GNNs), and graph transformers.

## 2.1 Supervised Learning methods for GRN Inference

**Supervised learning** is a fundamental approach in machine learning where algorithms are trained on labeled datasets—that is, datasets in which each input is paired with a known output. By analyzing these labeled examples, the algorithm learns to recognize patterns and relationships between inputs and their corresponding outputs. Once trained, the model can generalize this knowledge to make predictions on new, unseen data [27]. In the context of **gene regulatory network (GRN) inference**, supervised learning enables the prediction of direct downstream targets of transcription factors by leveraging labeled datasets containing experimentally validated regulatory interactions. This approach allows models to learn from known gene-regulatory relationships and apply that knowledge to uncover novel interactions with improved accuracy [28].

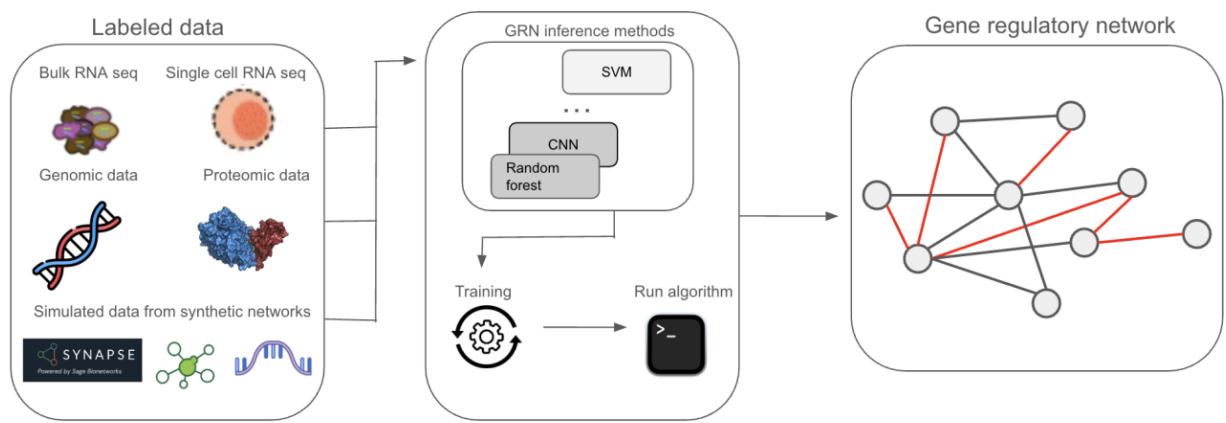

**Figure 2.** The process of training and testing supervised learning methods to infer GRNs.

**Figure 2** shows a typical process of training and testing supervised learning methods in the context of GRN inference. There have been different types of supervised methods developed for GRN inference, most commonly including random forest, support vector machine, regression-based methods, neural networks, and deep learning methods. Even though some prior works have compared and benchmarked various supervised methods with other methods and claimed that supervised methods perform better than other methods in some context[29][30] when sufficient labeled data are available, there are still significant challenges of applying supervised learning methods to most situations when no or few labeled data are available. In this section, we introduce the commonly used popular supervised learning algorithms in GRN inference and summarize the recent advances in supervised learning methods like supervised deep neural network architectures.

One of the most common methods of supervised learning algorithm, which is used in GRN inference, is *Random Forest*. GENIE3 is one of the popular algorithms that use the ensemble tree Random Forest to infer the GRN of *Escherichia coli* from high throughput microarray gene expression data [31]. GENIE3 breaks down the forecast of a gene regulatory network into various regression tasks and leverages Extra-Trees or Random Forest to predict the expression profile of a particular gene based on the expression profiles of all other genes. When predicting the target gene's expression pattern, GENIE3 takes into account the importance of an input gene as a possible regulatory connection. It generates a ranking of interactions that can be used to reconstruct the entire network by aggregating these putative regulatory links across all genes. It has performed well on both synthetic and real gene expression data and is the best performer in the DREAM4 *In Silico Multifactorial* challenge. GENIE3 is considered one of the state-of-the-art classic machine learning methods for GRN inference.

dynGENIE [32] is a semi-parametric version of GENIE3, which enables the analysis of both time-series data and steady-state data together. It employs Ordinary Differential Equations (ODE) to understand the expression of each gene over time and uses a non-parametric random forest model to learn the transcription function of each ODE. Inspired by the success of GENIE3, many adaptations and extended models of GENIE3 are developed. GENIMS is one of the extensions of GENIE3, which uses the guided random forest to solve individual regression problems and q-norm normalized for comparison, which are then refined to elevate the performance via variance of normalized weights as a guide [33]. BTNET[34] is an extension of GENIE3-time[35], which is a time-lagged version of GENIE3. The boosted tree used in BTNET differs from the bagging-based tree applied in GENIE3-time in that while GENIE3-time aggregates multiple independent estimators for constructing the final ensemble method considering the regulatory interactions between genes at different time points, the boosted tree continuously updates the estimator itself to make it stronger by compensating for the weakness of previous estimators. The other tools that incorporate random forest methods are iRafNET[36], GENREF[37], and GRRFNet[38]. TIGRESS [39] is another popular early algorithm that uses the random forest method to infer GRNs like GENIE3, however, the latter outperformed the former.

Another popular supervised learning algorithm is the *Support Vector Machine* (*SVM*). Support Vector Machine is considered an effective algorithm for solving general supervised binary classification problems [40]. In the context of gene regulatory interaction inference, SVM uses a kernel function as the main ingredient to measure the similarity between genes and finds the hyperplane that best separates the two classes of data points in the feature space [41]. SIRENE is one of the pioneer algorithms to use SVM in GRN inference leveraging gene expression data and known regulation relationships between transcription factors and target genes [42]. The method decomposes the problem of GRN inference into a large number of local binary classification problems and trains as an SVM binary classifier to discriminate between genes that are regulated and not regulated by a transcription factor based on gene expression data. The rationale behind this approach is that if two genes are regulated by the same transcription factor, then they are likely to exhibit similar expression patterns. SIRENE performed better than other unsupervised methods in some experiments [43].

CompareSVM is another software tool that predicts gene regulatory networks from expression data using a Support Vector Machine. It optimizes parameters, compares kernel accuracy by generating AUC (Area Under the Curve) for each kernel, and selects the best-performing kernel method for prediction [43].

One of the recent methods, the GRADIS approach, uses support vector machines to reconstruct gene-regulatory networks by providing feature vectors based on graph distance profiles from a network representation of the gene expression data. It was shown to outperform existing supervised approaches on the synthetic data and two benchmark datasets of Escherichia coli and Saccharomyces cerevisiae provided by the DREAM4 and DREAM5 network inference challenges [44]. Many other methods use supervised learning that uses the SVM approach, like Beacon GRN [45], and supervised ensemble approaches like EnGRaiN[46] for GRN inference.

RSNET is another supervised learning method that uses an information constraint-based approach to infer gene regulatory networks. It constrains candidate genes with highly dependent parameters measured from the data by mutual information (MI) as network enhancement items and highly putative candidate regulators as supervisors to improve optimization efficiency [47]. A lot of GRN inference tools also are developed on regression-based methods like Least Square Cut-Off (LSCO)[48], LASSO[49], and Ridge-regression with Cut-Off (RidgeCO) [50]. Least Square Cut-Off with Normalization (LSCON) [51] is one of the latest methods that is built on LSCO [52].

Recent advances in supervised deep learning have created significant waves in the field of technology and are now being implemented in various domains, including GRN inference. One of the most popular supervised neural network architectures is the convolutional neural network (CNN) [53]. DeepIMAGER[54] uses the ResNet50 CNN to infer GRNs and employs a supervised approach that converts the co-expression patterns of gene pairs into an image-like representation while incorporating improved transcription factor (TF) binding information for training. The dataset used in the study comprises single-cell RNA-seq (scRNA-seq) and ChIP-seq data, which capture TF–gene pair information across different cell types. It was shown that DeepIMAGER outperforms existing methods such as GENIE3, PIDC, SCODE, PPCOR, and SINCERITIES in some experiments.

Another supervised deep learning method for GRN inference is SPREd. SPREd[55] is a simulation-supervised neural network whose data include expression relationships among targets and between TFs within TF pairs. The model is trained using synthetic gene expression data produced by a simulation framework inspired by biophysical principles. This framework integrates both linear and nonlinear TF–gene interactions and simulates various GRN architectures. It was shown that SPREd performs better than other state-of-the-art models, such as GENIE3, ENNET, PORTIA, and others in some experiments, particularly on datasets with strong co-expression among TFs.

Recently, a cutting-edge deep learning technology, transformer, has emerged as a powerful deep learning architecture, particularly for modeling complex interactions among genes and transcription factors.

Originally designed for natural language processing tasks, the self-attention mechanism in transformers enables them to capture long-range dependencies, leading to greater efficiency for GRN inference—especially when gene interactions are complex and span multiple regulatory layers. When applied to GRNs, transformers can model how one gene regulates another across various time points or conditions, even when these relationships are nonlinear or occur over long distances. A key benefit of using transformers is their ability to work well with high-dimensional datasets, such as transcriptomic data, and to accurately construct gene interaction networks. For instance, research has applied transformer-based models to gene expression data, demonstrating that these models outperform traditional methods by capturing sequential and structural relationships within GRNs[56].

One tool that employs transformers is STGRNs. STGRNs consist of four components: the GEM module, the positional encoding layer, the transformer encoder, and the classification layer[57]. The GEM module converts gene pairs into a format suitable for input into the transformer encoder. The positional encoding layer extracts positional or temporal information. The transformer encoder computes the relationships among various sub-vectors, and the classification layer then makes the final categorization of the outcome. Results obtained with scRNA-seq data indicate that STGRNs outperform comparable tools in some experiments and are more interpretable.

DeepMAPS[58] is another application of transformers in the field of biological networks inference, which is related to the GRN inference. DeepMAPS is a deep learning network that leverages a heterogeneous graph transformer (HGT) framework to analyze single-cell data and infer cell-type-specific biological networks from scMulti-omics data (single-cell Multiomics Data), which enables the identification of active biological networks in different cell types and their response to external stimuli . Inferring biological networks involves identifying relationships between different genes or other biological entities, such as regulatory interactions or co-expression patterns. DeepMAPS is an end-to-end framework that takes in preprocessed single-cell multi-omics data and outputs inferred biological networks for each cell type. The framework consists of five major steps. (1) Data preprocessing: This step involves removing low-quality cells and lowly-expressed genes from the data. Different normalization methods are then applied based on the specific data types. An integrated cell-gene matrix is generated to represent the combined activity of each gene in each cell. Different data integration methods are applied for different scMulti-omics data types. (2) Building a heterogeneous graph: In this step, a heterogeneous graph is built from the integrated matrix, including cells and genes as nodes and the existence of genes in cells as edges. (3) Heterogeneous graph transformer (HGT) model building: An HGT model is built to jointly learn the low-dimensional embedding for cells and genes and generate an attention score to indicate the importance of a gene to a cell. (4) Predicting cell clusters and functional gene modules based on learned embeddings and attention scores: The HGT model generates an attention score to indicate the importance of a gene to a cell, which can be used to predict cell clusters and functional gene modules. (5) Cell-type-specific biological network inference: The HGT model enables the inference of cell-type-specific biological networks from scMulti-omics data.

The attention scores learned by the HGT model indicate the importance of each gene to each cell, which can be used to construct reliable gene association networks for each cell type. Autoencoders are used in DeepMAPS to generate initial embeddings for genes and cells. The architecture of the autoencoder differs depending on the type of data being used. Autoencoders are a type of neural network that can learn to compress and decompress data[59], which makes them useful for generating low-dimensional embeddings of high-dimensional data like single-cell multi-omics data. In DeepMAPS, two-layer graph neural network (GNN) autoencoders are used to generate initial embeddings for genes and cells from the integrated cell-gene matrix. The initial embeddings generated by the autoencoder are then updated using a heterogeneous graph transformer (HGT) model, which jointly learns the low-dimensional embeddings for cells and genes and generates an attention score to indicate the importance of a gene to a cell.

To infer GRNs, DeepMAPS first identifies basic gene regulatory modules (i.e., regulons) using three public functional databases: Reactome, Dorothea, and TRUST v2. Then, to avoid any bias in comparison, DeepMAPS compares cell-type-specific GRNs inferred from DeepMAPS with other state-of-the-art methods such as IRIS3[60] on scRNA-seq matrix or scATAC-seq matrix or both. The results show that DeepMAPS can accurately infer cell-type-specific GRNs with high interpretability. Furthermore, the inferred GRNs can be used to identify key regulators of specific biological processes or pathways in a given cell type. DeepMAPS is evaluated using benchmarking results that demonstrate its superior performance in cell clustering and biological network inference from scMulti-omics data.

However, there are some limitations with DeepMAPS. One of the main limitations is the computational efficiency for super-large datasets, which may contain billions of edges in the heterogeneous graph representation. This can be a practical issue, especially for datasets with more than 1 million cells. Additionally, DeepMAPS is recommended to be run on GPUs, which may lead to a potential problem of reproducibility since different GPU models have different floating-point numbers that may influence the precision of loss functions during the training process. As a result, DeepMAPS may generate slightly different cell clustering and network results for different GPU models. Another limitation is that DeepMAPS requires high-quality scMulti-omics data with low noise and batch effects to achieve optimal performance. In cases where the data quality is poor, or there are batch effects, DeepMAPS may not perform as well as expected.

One of our latest works, GRNFormer[61] is a deep learning network that leverages a variational graph transformer autoencoder to analyze single-cell RNA-seq data and infer gene regulatory networks (GRNs), thereby enabling the discovery of generalized regulatory interactions across diverse biological contexts and species. Inferring GRNs involves identifying relationships—such as co-expression patterns and transcription factor (TF) influences—between genes. GRNFormer is an end-to-end framework that takes preprocessed single-cell RNA-seq data as input and outputs a probabilistic gene regulatory network across cell types and species.

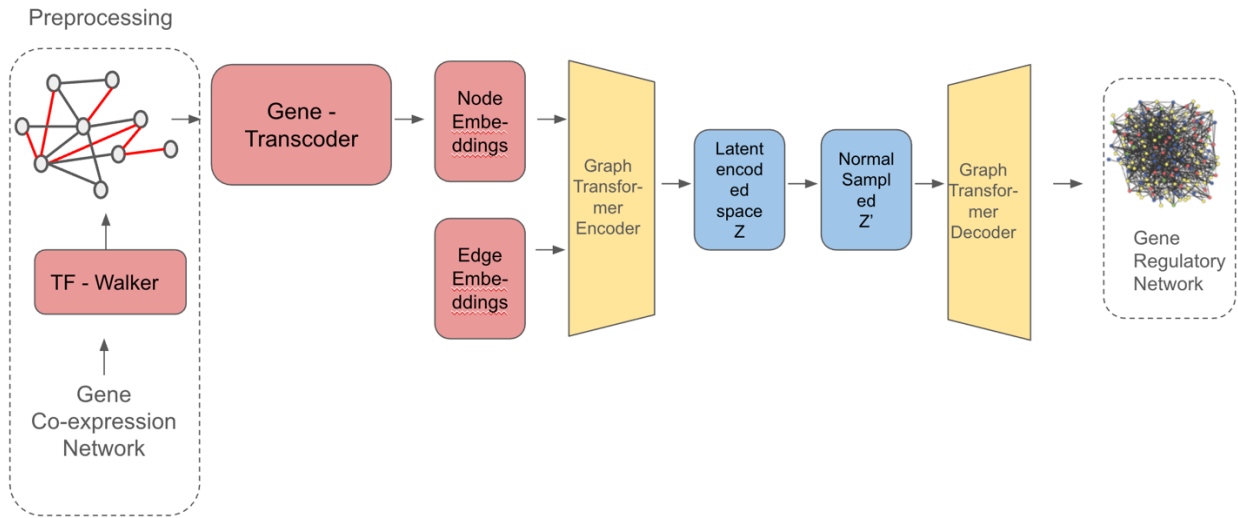

**Figure 3.** The architecture of GRNFormer

**Figure 3**. show the pipeline process of GRNFormer which begins with the construction of a gene co-expression network (GCEN). GRNFormer starts by normalizing gene expression data using ArcSinh, then constructs a gene co-expression network (GCEN) based on Pearson correlation, keeping only significant gene-gene associations. To handle high dimensionality, it uses a **TF-Walker** algorithm, which samples subgraphs centered on transcription factors by selecting nearby genes until a fixed size (100 nodes) is reached. Z-score normalization is applied within each subgraph to standardize expression ensuring that local expression contexts are accurately captured while reducing computational burden.

After subgraphs are sampled, GRNFormer processes them using the **GENE-Transcoder**, a transformer-based encoder. First, a 1D convolution captures local expression patterns from the cell-gene matrix. Then, transformer encoder layers with multi-head attention model both local and global gene interactions. A mean pooling operation generates compact, context-aware gene embeddings. These embeddings, along with GCEN edge features, are passed into a **variational graph transformer**

**autoencoder**, where multiple GRNFormer blocks integrate node and edge features. The encoder captures local and global regulatory patterns by computing pairwise attention scores and outputs latent representations, accounting for uncertainty in single-cell data through a Gaussian latent distribution. The GRNFormer decoder reconstructs regulatory interactions within subgraphs using additional GRNFormer blocks and a lightweight MLP. It refines node embeddings and generates edge attention scores, which are used to create a probabilistic adjacency matrix through an inner product and sigmoid activation, representing the likelihood of regulatory interactions. Predicted sub-networks are aggregated to form the full GRN. Training involves using ground truth regulatory data from sources like ChIP-seq and STRING networks, with a composite loss function combining binary cross-entropy and Kullback-Leibler (KL) divergence. GRNFormer is trained in a supervised manner, addressing class imbalance with dynamic negative sampling. This pipeline enables GRNFormer to infer accurate, cell-type-specific GRNs with strong performance and interpretability across diverse single-cell datasets.

GRNFormer was evaluated on several datasets across seven cell types and three regulatory networks (cell-type-specific ChIP-seq, non-specific ChIP-seq, and STRING). It achieved strong performance with an average AUPRC of 0.89 for non-specific networks, 0.83 for cell-type-specific networks, and 0.88 for STRING networks, showing its ability to capture gene regulatory relationships effectively. Notably, it excelled in generalizing to unseen cell types (mESC, mHSC-L), reaching AUPRC scores up to 0.95 and AUROC scores exceeding 92% across all datasets. In comparison to several popular GRN inference methods, GRNFormer outperformed all of them, achieving the highest AUPRC (up to 0.95) in non-specific networks and excelling in cell-type-specific and STRING networks. Its superior generalization and precision make it a powerful tool for GRN inference across diverse biological contexts.

The methodology of AnomalGRN[62] begins by reinterpreting GRN inference as a graph anomaly detection problem rather than a traditional pairwise link prediction task. In this approach, instead of considering each gene or gene pair individually, the method transforms gene pairs into nodes within a constructed graph. Each node is endowed with a feature vector that encapsulates combined gene expression profiles and associated regulatory signals. This transformation addresses the common imbalance in GRN inference—where true regulatory interactions are rare compared to a vast number of non-regulatory pairs—by enabling the focus on anomalous nodes that may represent true interactions. Central to the method is the use of a cosine metric rule to measure similarity between nodes. By evaluating the cosine similarity scores, the model distinguishes between groups of gene pairs that exhibit similar regulatory patterns (homogeneous) and those that deviate from these patterns (heterogeneous). The heterogeneous nodes, identified as anomalies, are posited to correspond to genuine regulatory relationships. To further refine the analysis, the model incorporates a graph sparsification technique. This step is designed to reduce the impact of noise and remove redundant or spurious connections that are common in single-cell RNA-seq data, thereby enhancing the clarity of the underlying network structure.

For the anomaly detection itself, the framework leverages advanced graph-based learning techniques that are adept at recognizing statistically significant deviations within the network. By systematically isolating these anomalous nodes, AnomalGRN can effectively predict regulatory interactions, overcoming the high dropout rates and technical noise that often obscure such signals in single-cell datasets. In terms of results, AnomalGRN was validated on multiple benchmark single-cell datasets, where it demonstrated robust performance in inferring GRNs. Quantitatively, the method achieved higher accuracy and stability compared to several state-of-the-art GRN inference methods. The model not only recapitulated known regulatory relationships but also uncovered novel hub genes and transcription factor–target interactions that had not been previously characterized. This suggests that by recasting the inference problem into an anomaly detection framework, AnomalGRN is particularly effective at identifying rare but biologically significant interactions amid a large background of noise.

## 2.2 Unsupervised Learning for GRN Inference

Unsupervised learning plays a crucial role in the inference of GRNs, particularly when labeled data are insufficient or unavailable. This approach enables scientists to uncover hidden patterns and structures within large-scale gene expression datasets, potentially revealing regulatory interactions that might otherwise go unnoticed. As the complexity and scale of biological data continue to expand, unsupervised learning methods have become increasingly essential for advancing our understanding of gene regulation.

Evolutionary machine learning (EML)[63] combines evolutionary algorithms with traditional machine learning to tackle complex optimization problems. Motivated by biological evolution, EML employs operators such as selection, mutation, and crossover to generate solutions across multiple generations. This method is useful for problems that require exploring large, multidimensional spaces, such as optimizing neural network architectures or fine-tuning hyperparameter. Furthermore, to provide an innovative and reliable consensus approach based on previous outcomes, GENECI (gene network consensus inference)[64] was developed. This tool uses an evolutionary machine learning strategy to organize clusters, compute the results of the main inference method, and optimize the network based on topological features and confidence rates. The tool has been tested in various experiments—including the DREAM challenge and the IRMA network—with results demonstrating robust outcomes, thereby supporting its potential application in clinical settings for melanoma treatment.

Information theory[65]—a unsupervised mathematical framework for quantifying the amount of information shared among variables—plays an important role in analyzing GRNs. It allows quantifying relationships among genes by measuring the dependency of one gene's expression on another. A key metric derived from information theory is mutual information (MI), which is widely used in network inference because it can detect both linear and nonlinear relationships among variables. This approach does not require extensive prior knowledge and can handle large numbers of genes, providing a significant advantage in capturing nonlinear relationships. However, applying MI to continuous data (e.g., normalized fluorescence intensity measurements of gene expression) can be influenced by factors such as sample size, correlation strength, and underlying data distributions. Consequently, optimizing MI calculations often involves complex adjustments and manual fine-tuning, which can sometimes be arbitrary and time-consuming.

Among the most well-known information theory approaches for GRN inference is ARACNE (Algorithm for the Reconstruction of Accurate Cellular Networks). ARACNE[66] is based on information theory and employs MI to identify statistical dependencies among gene expression profiles. By quantifying the reduction in uncertainty in one gene's expression given another, MI serves as a useful metric for detecting potential regulatory links. However, the direct application of MI often results in the detection of indirect interactions that may not be biologically relevant. To address this issue, ARACNE applies the data processing inequality (DPI) to eliminate these indirect interactions, thereby producing a clearer and more biologically precise network—especially useful in large-scale analyses where data complexity can introduce significant noise

Similarly, MRNET (Minimum Redundancy Networks)[67] is a tool that incorporates the concept of reducing redundancy in feature selection for GRN inference. MRNET selects the most informative features while minimizing redundancy, which is critical in biological datasets where repetitive information can obscure true regulatory signals. By focusing on the most relevant and distinct features, MRNET enhances the precision and clarity of the inferred networks, making it an essential approach for researchers working with noisy data in high-dimensional spaces that might otherwise lead to overfitting. Evaluations based on thirty synthetic microarray datasets indicate that MRNET performs competitively. Furthermore, CLR (Context Likelihood of Relatedness) builds upon ARACNE's foundation by incorporating an additional layer of contextual analysis. While ARACNE relies primarily on MI, CLR examines the background distribution of MI scores within the dataset to highlight interactions that stand out from the noise. By contextualizing each interaction, CLR improves the reliability of network inference, making it particularly effective in datasets with high variability and noise—common in complex biological systems.

During the last several years, several unsupervised deep learning methods were developed to infer GRNs. For instance, bidirectional recurrent neural networks (Bidirectional RNNs) have been employed in GRN inference to analyze sequential gene expression data. Unlike traditional (unidirectional) RNNs—which only incorporate past information—bidirectional RNNs process data in both forward and backward directions. This dual approach enables them to capture dependencies from both past and future time points, making them particularly well-suited for modeling time-series gene expression data where the order of gene activation is critical. Consequently, bidirectional RNNs[68] can identify regulatory relationships that span multiple time points, enhancing prediction accuracy BiRGRN (bidirectional recurrent gene regulatory networks)[69] is a tool that combines deep learning with an unsupervised framework. BiRGRN employs bidirectional RNNs to analyze time-series gene expression data, effectively capturing temporal dynamics that are essential for understanding gene regulation. In contrast, standard RNNs—limited by their unidirectional processing—can miss important temporal dependencies. By processing data in both directions, BiRGRN offers a more comprehensive view of regulatory interactions. This capability is especially useful in studies of dynamic biological processes, such as development and responses to environmental stimuli, where precise timing of gene expression plays a vital role. Experiments on four simulated datasets and three real scRNA-seq datasets demonstrate that BiRGRN can simultaneously infer GRNs from time-series scRNA-seq data, outperforming some existing methods.

Another widely applied unsupervised learning algorithm is the variational autoencoder (VAE). VAEs[70] are generative models that have been used to interpret GRNs by analyzing the latent representations of gene expression data. They enable dimensionality reduction, which makes it easier to discover underlying patterns in high-dimensional gene expression datasets. By mapping gene expression data to a lower-dimensional latent space, VAEs can reveal hidden networks by reconstructing the original expression data from these latent variables. GRN-VAE[71] is an advanced approach that leverages the power of VAEs to uncover complex, nonlinear regulatory relationships within gene expression data. VAEs compress data into a low-dimensional representation while preserving critical details. GRN-VAE uses this compressed representation to infer subtle, nonlinear regulatory interactions that traditional linear methods might miss. By capturing these complex dependencies, GRN-VAE offers a clearer, more nuanced understanding of gene regulation—particularly in high-dimensional datasets typical of higher organisms, where multiple layers of regulation interact.

Additionally, DeepSEM (deep structural equation modeling)[72] represents a hybrid approach that combines the strengths of structural equation modeling (SEM) with deep learning methods. SEM is a statistical technique typically used to model relationships between observed and latent variables, which makes it suitable for GRN inference when direct relationships among genes are not readily observable.

DeepSEM extends this framework by incorporating deep learning to capture both linear and nonlinear dependencies, resulting in a flexible and robust method for inferring GRNs in complex biological systems.

Recently, another deep learning designed for graph-structured datasets called graph neural networks (GNNs) was also applied to GRN inference. Unlike traditional neural networks, which process Euclidean data such as images or text, GNNs excel at analyzing non-Euclidean data, where relationships among entities are represented as nodes connected by edges in a graph. GNNs enhance both the representation of individual node features and the overall structure of graphs, making them well-suited for applications such as social network analysis, molecular structure prediction, and GRN inferenc[73]. For example, CVGAE[74] applies a graph neural network that combines gene expression data with network topology to embed the data into a low-dimensional vector space. This vector is then used to compute distances between genes and predict interactions. CVGAE employs multi-stacked GraphSAGE layers as the encoder and an enhanced decoder to address network sparsity. Evaluations on various single-cell datasets—including four ground-truth networks—indicate that CVGAE performs exceptionally well compared with other tools.

Despite the significant progresses made by unsupervised learning methods in GRN inference, they also have limitations. For instance, because they are not trained with labels (ground truth GRN networks), they are susceptible to inferring spurious correlations between genes instead of causal regulatory relationships. Therefore, there is a need to leverage label information, if available, with unsupervised GRN inference methods.

## 2.3 Semi-Supervised Learning for GRN Inference

Semi-supervised learning occupies a unique middle ground between supervised and unsupervised learning by combining both labeled and unlabeled data. This method is particularly useful in biological studies where obtaining labeled data can be expensive and time-consuming, while large amounts of unlabeled data are readily available. In the context of GRNs, semi-supervised learning leverages limited experimental data labels along with abundant unlabeled gene expression data, thereby enhancing the precision and generalization of the inferred networks.

TSNI (Time Series Network Inference)[75] is a semi-supervised tool for GRN inference from time-series gene expression data. TSNI is especially well-suited for inferring dynamic regulatory interactions that are critical for understanding processes such as cell differentiation, circadian rhythm, and responses to environmental changes. The approach uses both current labeled data (reflecting existing regulatory interactions) and unlabeled time-series data to predict how genes influence each other over time. By refining a dynamic system model, TSNI captures temporal dependencies and infers direct causal relationships between genes. This method is particularly effective for studies where the evolution of gene expression is essential for understanding the regulatory network.

Genetic algorithms (GAs)[76] are optimization techniques inspired by natural selection, where candidate solutions evolve over generations to approach an optimal solution. In GRN inference, GAs can be applied to optimize the structure of the regulatory network by evolving subsets of gene interactions. The semi-supervised component is introduced by using labeled data to guide the evolutionary process, while the unlabeled data provide additional context to fine-tune the network. Fixed subsets of labeled data serve as constraints during optimization, ensuring that the inferred network remains biologically plausible (Larranaga et al., 1999). In [77], the authors proposed a genetic algorithm for fixed-size subset selection, combined with an SVM to enhance performance. Their experiments on both simulated and real-world datasets demonstrate that the proposed algorithm performed well when the data are balanced—and successfully identifies optimized solutions for each transcription factor examined in the study.

Finally, an advanced semi-supervised deep learning method for GRN inference is GRGNN (Graph Recurrent Gene Neural Networks)[78], which combines the strengths of graph neural networks (GNNs) and recurrent neural networks (RNNs) to model both the graph structure of GRNs and the temporal dynamics in gene expression data. GRGNN constructs a graph where nodes represent genes and edges represent potential regulatory interactions. It then analyzes both the graph structure and the dynamic gene expression data using a mix of labeled and unlabeled data. The semi-supervised nature of GRGNN allows it to use a small set of experimentally verified regulatory relationships to guide the inference process, while refining the network structure using the unlabeled data. This approach is especially useful when modeling GRNs with complex topological features, where the expression of each gene is influenced by its neighbors.

**2.4 Contrastive Learning for GRN Inference**

**Contrastive learning** is a powerful *self-supervised* paradigm that has recently been harnessed for GRN inference. The key idea is to train models to map high-dimensional biological inputs into a latent embedding space where samples that share regulatory similarities are positioned close together while those with dissimilar regulatory roles are pushed apart. This is typically achieved by constructing positive pairs—such as different augmented views of the same gene pair—and negative pairs—such as gene pairs without known regulatory interactions—and optimizing a contrastive loss function like the Information Noise-Contrastive Estimation (InfoNCE) loss [79][80]. The InfoNCE loss encourages high similarity between positive pairs and imposes a penalty for similarities among negative pairs, often using techniques like temperature scaling to refine the learning dynamics[81].

DeepMCL[82] is a deep multi-view contrastive learning model developed to infer gene regulatory networks from single-cell RNA-sequencing data obtained from diverse platforms and time points. It overcomes challenges associated with noisy measurements and dropout events typical in single-cell data. In this framework, each gene pair is represented as a histogram image that captures co-expression relationships by dividing expression levels into distinct bins. A Siamese convolutional neural network extracts low-dimensional embeddings from these images using contrastive learning to distinguish between positive gene pairs, which have regulatory interactions, and negative pairs lacking such interactions. Additionally, an attention mechanism integrates embeddings from multiple data sources and neighboring gene pairs, thereby emphasizing informative features and reducing false positives. The architecture employs a VGG-style convolutional neural network and incorporates non-local blocks to expand the receptive field and enhance feature extraction. The model operates in two stages: feature extraction through contrastive loss and subsequent regulatory interaction prediction using concatenated embeddings and fully connected layers. Experimental evaluations on synthetic and real-world datasets demonstrate that DeepMCL effectively integrates multi-view data to accurately predict transcription factor–gene interactions while mitigating noise and redundancy inherent in single-cell measurements. This innovative framework deepens insights into cellular regulation and gene interaction dynamics with good accuracy.

Another approach, GCLink—Graph Contrastive Link Prediction Framework—[83] extends contrastive learning to the link prediction task within GRNs. Unlike methods that consider gene pairs in isolation, GCLink leverages the entire network structure by learning low-dimensional embeddings that reflect both local and global topological properties. In GCLink, positive examples are drawn from observed regulatory links, while negative examples are sampled from unconnected gene pairs. Graph augmentation techniques, such as node dropout or edge perturbation, are used to generate diverse views of the network, which in turn produce robust training pairs. The contrastive loss in GCLink ensures that the embeddings of gene pairs with known regulatory interactions are aligned closely, while embeddings of unrelated pairs are separated by a defined margin. This framework not only improves the resolution of GRN reconstruction but also enhances robustness against noise by fully leveraging the inherent structural properties of the graph.

Together, these contrastive learning frameworks—DGCRL and GCLink—demonstrate that integrating contrastive objectives into GRN inference can significantly improve the detection and characterization of regulatory interactions, offering a robust alternative to traditional methods in the face of complex, noisy biological data.

## 3. Types of Inputs and Outputs of GRN Inference

## 3.1 Types of Outputs of GRN inference

GRN Inference methods can be classified into groups according to the output that they produced: (1) Local GRN inference methods and (2) Global GRN inference methods.

**Local GRN Inference.** Local GRN inference focuses on identifying regulatory relationships for a specific gene or a small group of genes. These methods typically examine direct interactions and employ statistical or machine learning techniques that perform well on smaller, high-quality datasets. For example, one common approach is to use linear or nonlinear regression models to determine how the expression level of a potential regulator affects that of a target gene. The Inferelator algorithm, for instance, uses a regression technique to infer sparse regulatory networks by identifying transcription factors that best predict the expression of target genes[84]. Another approach relies on mutual information; the MRNET algorithm measures the mutual dependence among gene expression profiles to detect regulatory relationships. While MRNET efficiently discovers direct regulatory interactions, it may not capture broader network contexts[85].

**Global GRN Inference.** Global GRN inference seeks to reconstruct the entire network of gene regulatory interactions within a cell or an organism. This comprehensive strategy considers both direct and indirect interactions across the genome, providing a holistic view of the regulatory landscape. Global inference methods are essential for understanding complex biological systems, such as developmental processes, disease mechanisms, and responses to environmental stimuli. These methods typically operate on high-dimensional data and must account for inherent noise in large datasets. For example, GENIE3 applies machine learning techniques—specifically, random forest—to rank potential regulatory interactions for each gene across the entire dataset[86]. Another model, ARACNE, uses mutual information combined with the data processing inequality to eliminate indirect interactions, thereby producing a more precise global network[87]. Global approaches utilize large-scale data types, such as bulk RNA-seq, single-cell RNA-seq, and datasets from projects like GTEx, to construct comprehensive GRNs.

After a GRN is inferred, it is useful to visualize it to facilitate biological knowledge discovery. Cytoscape[88] is a widely used open-source software platform for visualizing and analyzing complex networks including GRNs, particularly in bioinformatics and systems biology. Its standout feature is its ability not only to display intricate molecular interaction networks but also to integrate rich attribute data—such as gene expression profiles, functional annotations, and more—directly onto the network. This integration provides a multidimensional view of the biological systems under study. One of the most user-friendly aspects of Cytoscape is its intuitive interface, which makes it easy to import network data from various sources. Once the data is loaded, users can customize the visualization by adjusting node sizes, colors, and layouts to highlight the most relevant features, thereby enhancing interpretability. This level of customization is invaluable for extracting insights from complex datasets. Another major strength of Cytoscape is its extensibility. It supports a vast ecosystem of plugins, or "apps," developed by both the core team and the broader community. These apps extend Cytoscape's functionality in numerous ways. For example, some apps focus on network topology analysis, helping users examine structural properties of their networks. Other apps facilitate clustering by grouping nodes based on specific criteria, and still others perform enrichment analyses to identify overrepresented functions or pathways within the network.

### 3.2 Types of Input Data for GRN Inference

GRN inference methods usually take some high-throughput omics data as input to infer GRNs. Different GRN inference methods may work with different types of data. The most commonly used data are genomics and transcriptomics data because of their near universal availability, while other omics data can also provide complementary information if available. Below are some major data sources that can be leveraged for GRN inference.

- **Genomic Data Sources:** Genomic data covers the complete DNA sequence of an organism, including genes, regulatory elements, and genetic variants (for example, SNPs). This information is essential for identifying regulatory regions and transcription factor (TF) binding site[89]. Large-scale projects like the 1000 Genomes Project and the International Cancer Genome Consortium (ICGC) also offer in-depth genomic and mutation data, illuminating how different genetic variants might influence GRNs.

- **Transcriptomic Data Sources:** Transcriptomic data captures all RNA transcripts under certain conditions. Methods like RNA sequencing (RNA-seq) quantify gene expression levels and drive GRN inference approaches such as ARACNe and GENIE3[90][91]. Public repositories, including the Gene Expression Omnibus (GEO), ArrayExpress, and the Genotype-Tissue Expression (GTEx) project, host extensive transcriptomic datasets for various tissues and populations, facilitating large-scale GRN reconstructions.

- **Epigenetic Data Sources:** Epigenetic modifications (e.g., DNA methylation and histone modifications) affect gene expression without altering the DNA sequence. Techniques such as ChIP-seq provide insights into protein–DNA interactions and chromatin states[92]. Consortia like ENCODE and Roadmap Epigenomics generate massive datasets covering diverse epigenomic marks, helping pinpoint active regulatory regions for GRN inference. However, epigenetic data is less available than genomics and transcriptomics data, and even it is available, many GRN inference methods still cannot leverage it.

- **Proteomic Data Sources:** Proteomic data focuses on the large-scale identification and quantification of proteins, revealing post-transcriptional regulation and protein–protein interactions [93] Mass spectrometry–based initiatives such as the Clinical Proteomic Tumor Analysis Consortium (CPTAC) produce proteomic and phosphoproteomic profiles linked to genomic data, shedding light on how changes at the protein level refine GRN models. Like epigenetic data, proteomic data is less available than genomics and transcriptomics data and less leveraged.

- **Single-Cell Multi-Omics Data.** This kind of data contains multiple omics data such as transcriptomics and ATAC-seq data. The single-cell methods can measure multiple molecular layers (e.g., RNA expression and chromatin accessibility) within individual cells, capturing cell-specific regulatory networks[94][95]. Integrated pipelines like Seurat[96] and MOFA[97] use these multi-omics data to refine GRN inferences by highlighting variability at the single-cell level.

- **Gene Expression Data + Protein Interaction Data.** Incorporating transcriptomic data with protein–protein interaction (PPI) networks strengthens GRN inference by adding physical interaction evidence. Databases like STRING and BioGRID consolidate comprehensive PPI data. Methods such as PANDA integrate these interaction networks with gene expression measurements, enabling the refinement of regulatory relationship predictions[98]

**Table 2**. A list of major data sources for GRN inference.

| Dataset Name | Omics Type | Source | Description |
|---|---|---|---|
| ENCODE | Epigenomic | NIH | Catalogs TF binding, chromatin marks, and regulatory elements. |
| TCGA | Multi-omics | NCI | Cancer genomic data (DNA, RNA, epigenetics, proteomics) for over 30 tumor types. |
| ICGC | Multi-omics | International Consortium | Global effort sequencing 50+ cancer types (genomic, transcriptomic, epigenomic). |
| GTEx | Transcriptomic/Genomic | NIH | Tissue-specific expression data linked with donor genotypes (eQTLs). |
| Roadmap Epigenomics | Epigenomic | NIH | Reference epigenomes (histone marks, DNA methylation) across diverse human tissues. |
| BLUEPRINT | Epigenomic | EU FP7 (BLUEPRINT) | Epigenomes of blood cells (histone marks, methylation). |
| GEO | Transcriptomic | NCBI | Repository for functional genomics data (RNA-seq, microarray). |
| ArrayExpress | Transcriptomic | EMBL-EBI | Functional genomics archive, overlapping with GEO. |
| CCLE | Multi-omics | Broad / Novartis | Data for 1000 cancer cell lines (genomics, expression, drug response). |
| LINCS L1000 | Transcriptomic | NIH LINCS / Broad | Large perturbation dataset (¿1M profiles) capturing expression changes. |
| Human Cell Atlas | Multi-omics | HCA Consortium | Single-cell data (RNA, ATAC) from various human tissues. |
| Cistrome DB | Epigenomic | X. Liu Lab | Curated ChIP-seq/ATAC-seq for TF binding and chromatin accessibility. |
| TRANSFAC | Genomic / Regulatory | geneXplain | TF binding motifs and consensus sites (license required). |
| Oncomine | Transcriptomic | Thermo Fisher | Cancer gene expression platform with curated datasets. |
| CPTAC | Proteomic | NCI | Proteomic (protein/phosphoprotein) data linked to TCGA tumor samples. |

**Table 2** lists a number of major sources where researchers can retrieve different types of data to infer GRNs in addition to using their in-house data. Public resources (ENCODE, TCGA, etc.) are freely accessible and provide the community with large-scale data to derive and validate regulatory networks. Commercial or restricted databases (like TRANSFAC and Oncomine) often aggregate insights from public data or literature in a convenient form, albeit behind paywalls; when available, they can complement open data by adding prior knowledge or extended datasets. Researchers typically use multiple data sources – for example, using ENCODE/epigenomic data to refine networks learned from expression data (GEO/TCGA), or leveraging multi-omics projects (like GTEx and Roadmap) to build condition-specific GRNs with stronger confidence. Combining these diverse datasets ultimately leads to more accurate and context-specific GRNs, as each data type provides a different "view" of the regulatory landscape.

Moreover, developing versatile GRN inference methods to use multiple sources of data whenever available is important for improving GRN inference because multiple complementary data can provide more insights into underlying gene regulatory mechanisms. However, integrating multiple modalities of data (e.g., multi-omics data) to infer GRNs is still a major challenge in the field.

**4. Gold Standard Datasets for Training and Testing GRN Inference Methods**

Obtaining enough high-quality labeled data is critical for training and/or testing machine learning methods to address any scientific problem, including the GRN inference. Below is a summary of the main datasets available for training and testing GRN inference methods.

1. **DREAM Bulk RNA-seq dataset**: Bulk RNA sequencing is a technique used to measure gene expression levels by sequencing RNA from a heterogeneous mixture of cells. This method provides an average expression profile for each gene across all cells in the sample, which can obscure cell-to-cell variability. Nonetheless, bulk RNA-seq is important for understanding general gene expression patterns in tissues. GRN models such as ARACNe and GENIE3 use bulk RNA-seq data from DREAM dataset (https://www.synapse.org/Synapse:syn3049712/wiki/74630) [99][100] to interpret gene regulatory networks by examining co-expression patterns and mutual information among genes.

2. **Single-cell RNA-seq (scRNA-seq) dataset**: scRNA-seq profiles gene expression at the individual cell level, allowing researchers to study cellular heterogeneity and identify rare cell populations. This high-resolution data is essential for constructing GRNs that capture cell-specific regulatory interactions. Tools such as SCENIC and PIDC utilize scRNA-seq from Zeisel et al 's dataset (http://www.ncbi.nlm.nih.gov/geo/query/acc.cgi?acc=GSE60361) [101][102][103] to infer regulatory networks by combining gene expression information with regulatory models.

3. **GTEx**: The Genotype-Tissue Expression (GTEx) project is a comprehensive resource that provides gene expression and regulatory data from various human tissues obtained from healthy individuals. It enables the study of tissue-specific gene expression and regulatory mechanisms. GRN models such as PANDA have used GTEx data (https://www.gtexportal.org/home/) to construct tissue-specific regulatory networks, thereby enhancing our understanding of gene regulation in different biological contexts[104][105]

4. **DREAM4 and DREAM5 Datasets**: The DREAM (Dialogue for Reverse Engineering Assessments and Methods) Competition offers benchmark datasets for evaluating GRN inference methods. DREAM4 and DREAM5 consist of synthetic and real gene expression data with known regulatory interactions, serving as excellent standards for testing algorithms such as GENIE3, TIGRESS, and Inferelator

5. **ChIP-seq data: ChIP-seq** (Chromatin Immunoprecipitation Sequencing) is a technique used to identify DNA binding sites for transcription factors and other DNA-associated proteins, providing direct evidence of regulatory interactions. GRN models often use ChIP-seq data to validate predicted interactions or incorporate them as prior knowledge.

6. **Multi-omic Single-cell Datasets**: In addition to scRNA-seq data, single-cell datasets can include modalities such as single-cell ATAC-seq and single-cell methylation profiles. These datasets enable multi-omics integration in GRN modeling, offering a more comprehensive view of gene regulation. Tools like scMTNI combine single-cell multi-omics data to infer regulatory networks and uncover hidden biological processes within Chen et al 's dataset (https://github.com/pinellolab/scATACbenchmarking/tree/master/Real_Data/Buenrostro_2018)[106]

7. GRNdb: GRNdb is a database that provides gene regulatory networks derived from RNA-seq data across various species, including humans, mice, and Arabidopsis. It serves as a resource for accessing precomputed GRNs and can be used to validate new network predictions. GRN models can use GRNdb as a reference for comparison and validation[107].

8. Reactome: Reactome is a curated database of pathways and reactions in human biology that offers detailed information on molecular processes. It informs GRN models by providing data on known biological pathways and interactions involved in gene regulation. Resources such as Pathway Commons and PSIA integrate Reactome data to enhance network inference and pathway analysis[108][109].

9. DoRothEA: DoRothEA is a comprehensive resource that compiles human and mouse transcription factor regulatory data with confidence levels. It is used alongside gene expression data to predict TF activity and interpret regulatory networks. For example, GRN models like VIPEP utilize DoRothEA to perform network analysis and estimate TF activity from gene expression profiles[110].

10. TRUST v2: TRUST v2 is a manually curated database of human and mouse transcriptional regulatory networks that provides detailed insights into TF-target interactions, including regulatory direction and mechanisms. GRN models can incorporate TRUST v2 data to validate inferred networks or use it as prior knowledge to guide network inference, thereby enhancing the biological relevance of their predictions[111].

11. **KEGG (Kyoto Encyclopedia of Genes and Genomes).** KEGG offers comprehensive pathway maps that detail molecular interactions and reactions, serving as a widely used reference for annotating genes and proteins [112].

12. **WikiPathways.** This community-maintained database features a diverse collection of curated biological pathways, making it a valuable resource for integrating pathway-level information into GRN inference [113].
13. **RegulonDB.** Focused on *Escherichia coli*, RegulonDB curates detailed information on transcriptional regulation, including binding sites and operon organization, which is essential for constructing accurate regulatory networks [114]

In addition to the GRN datasets above, there is a tool GeneNetWeaver (GNW)[115] that can generate synthetic gene expression data based on known network topologies. It is widely used in the DREAM challenges to create datasets with known regulatory networks, which are then used to benchmark GRN inference algorithms such as ARACNe, GENIE3, and CLR. However, it is worth noting that, even though it is useful to train and test GRN inference methods on simulated data, it is still necessary to test them on real-world data to assess how well they work.

## 5. Evaluation Metrics for GRN Inference

Evaluating predicted GRNs is crucial for assessing the accuracy and reliability of inference methods. It helps identify their strengths and weaknesses, providing insights into performance across different contexts. Validation and benchmarking ensure the correctness and robustness of reconstructed GRNs, guiding researchers in selecting the most reliable methods for their studies. This prevents inaccurate reconstructions that could lead to false conclusions or predictions. Below is a list of common metrics for evaluating inferred GRNs.

### 5.1 Common evaluation metrics

**AUROC** stands for the Area Under the Receiver Operating Characteristic Curve. It is a metric used to evaluate the accuracy of classification models, such as those used to infer GRNs. AUROC is calculated by plotting the true positive rate (TPR) against the false positive rate (FPR) at different threshold values for the predicted edges (gene regulatory interactions). The area under the resulting curve is then calculated to obtain the AUROC value, which ranges from 0 to 1, with higher values indicating better performance.

**AUPRC** stands for Area Under the Precision-Recall Curve. It is also a common metric used to evaluate the accuracy of inferred GRNs. However, AUPRC is calculated by plotting the precision against the recall at different threshold values for the predicted gene regulatory interactions. The area under the resulting curve is then calculated to obtain the AUPRC value, which ranges from 0 to 1. AUPR concentrates more on performance in searching for true positives than AUROC while reducing false positives, making it more informative in GRN inference tasks in comparison to AUROC.

**Precision** and **recall** provide granular insights into model performance at a specific threshold. Precision measures the proportion of predicted regulatory interactions that are correct (TP / (TP + FP)), while recall assesses the proportion of actual regulatory interactions that are successfully identified (TP / (TP + FN)). Here TP, FP, and FN denote the number of true positives, false positives, and false negatives, respectively. In GRN inference, there is typically a trade-off between these two metrics: increasing recall can result in more false positives and thus lower precision, and vice versa. These metrics are essential for understanding whether a model favors sensitivity (recall) over specificity (precision) or achieves a balance between the two.

**F1 score** is the geometric mean of precision and recall (i.e., 2 × precision × recall / (precision + recall)), which combines precision and recall into a single metric that balances the trade-off between the two. It provides a more comprehensive perspective on model accuracy when both metrics are critical. This is particularly useful in GRN inference, where precision and recall are often equally important.

## 5.2 Evaluation Framework and Benchmark

Rigorously and objectively evaluating GRN inference methods is important and challenging. Standard evaluation frameworks and benchmarks can streamline the evaluation process and make it easier for users and developers alike to evaluate GRN inference methods. BEELINE is a systematic framework developed to evaluate the accuracy of the methods that infer GRNs from single-cell gene expression data [116]. It uses synthetic networks with predictable cellular trajectories as well as curated Boolean models to serve as the ground truth for evaluating the accuracy of GRN inference algorithms. BEELINE aids in evaluating GRNs by providing a strategy to simulate single-cell gene expression data from these two types of networks that avoid the pitfalls of previously used methods. The framework also provides recommendations to users of GRN inference algorithms, including suggestions on how to create simulated gene expression datasets for testing them. BEELINE is available at http://github.com/murali-group/BEELINE under an open-source license and will aid in the future development of GRN inference algorithms for single-cell transcriptomic data.

## 6. Challenges and Future Direction

Despite the significant progress made by machine learning methods above, there are several limitations and challenges in the field of GRN inference. The first major challenge is that there is a lack of standard method (like AlphaFold[117] for protein structure prediction) that can generally make high-accuracy GRN inference for different cells and different species in different biological conditions. A tool can only reasonably capture one or a few aspects of a GRN for some cells in some conditions. No method can always outperform others in inferring putative transcriptional targets, putative post-translational targets, or master regulators that drive certain phenotypes [118]. Therefore, it is important to develop sophisticated AI methods that can generalize well to all kinds of real-world biological environment. Mimicking how deep learning has revolutionized protein structure prediction, one direction is to develop more sophisticated deep learning methods such as transformers and diffusion models[119] that are suitable for representing multiple sources of omics data and the interactions between them to accurately infer GRNs in different biological contexts, regardless of species and cells. Simply applying an off-shelf deep learning method to GRN inference will unlikely yield optimal results. The advanced deep learning methods specially customized for GRN inference like AlphaFold2 and AlphaFold3 specially designed for protein sequence and structures are needed to improve the accuracy of GRN inference across the board.

An emerging avenue in addressing these challenges is the integration of foundation models into GRN inference. Foundation models, which are large pre-trained neural networks that capture broad representations from massive datasets, have demonstrated exceptional performance in natural language processing and computer vision[120]. By fine-tuning such models on domain-specific data have shown that they can effectively extract meaningful biological insights even from complex omics datasets. In the context of GRN inference, foundation models could be adapted to learn representations that capture the intricate relationships among genes, transcription factors, and regulatory elements. This approach not only leverages vast amounts of heterogeneous data but also allows for more flexible model-based inference, where the model's learned representations can be used as priors to improve the inference of regulatory networks. The development of such models promises to mitigate issues related to data sparsity and heterogeneity while providing uncertainty estimates that enhance the reliability of the inferred networks.

The second major challenge is to integrate multi-omics data, particularly, increasingly popular single-cell multi-omics (sc-Multi-omics) data that is very sparse and of high dimensionality, which makes it difficult to identify meaningful patterns and relationships between genes. Another difficulty is the heterogeneity of scMulti-omics data, which may contain different types of cells with distinct gene expression profiles and regulatory mechanisms. Furthermore, GRN inference from scMulti-omics data requires the integration of multiple types of omics data, such as scRNA-seq and scATAC-seq, which may have different levels of noise and bias. Integrating these different types of omics data is difficult due to technical limitations and differences in experimental protocols. Moreover, there is a need for accurate cell clustering to identify cell-type-specific gene expression profiles and regulatory mechanisms. However, accurate cell clustering can be difficult to achieve due to noise, batch effects, and other confounding factors in scMulti-omics data. Therefore, GRN inference from scMulti-omics data requires advanced and robust AI methods that can handle large-scale datasets with high dimensionality and complexity, addressing the issues related to data quality, heterogeneity, integration, cell clustering accuracy, and computational efficiency [121]. This call for the development of more innovative AI methods, particularly deep learning models like multi-modal AI models for text, image and video processing, to tackle this challenge.

The third major challenge is the lack of reliable real ground-truth GRNs against which to training and evaluate GRN inference methods. Despite there are some ground-truth networks available (see Section 4), the amount of data is still very limited and not sufficient to train GRN inference methods that can generalize well to different biological conditions, considering the complexity of GRN inference. Moreover, the existing ground-truth networks are usually incomplete and miss many regulatory interactions, making it hard to train and test GRN inference methods. Due to this problem, simulated data have been widely used to assess the performance of network inference methods. However, these simulated data sets may not always accurately represent the real-world gene regulatory networks [122] and cannot substitute the real-world GRN data. One way to tackle this challenge is to extract more ground-truth gene regulatory networks from biomedical literature. Sophisticated large language models (LLMs) such as ChatGPT may be able to help automate this process to some degree upon well-designed prompts. Therefore, how to design prompts for LLMs to accurately retrieve known GRNs buried in the literature can be an interesting direction to pursue. Moreover, creating a central database to store all the known GRNs and the corresponding input data like the Protein Data Bank (PDB) for protein structure is also important to enable the machine learning and AI community to develop sophisticated GRN inference methods. Future research can explore prompt design and fine-tuning strategies for LLMs to accurately retrieve and integrate known GRNs from the literature, ultimately contributing to the creation of such a central, comprehensive.

Finally, most existing methods focus on inferring static GRNs, even though GRNs dynamic changes in cells in response to internal and external stimuli. It is still very challenging to infer dynamic GRNs [123]. Current methods lack flexibility when it comes to specifying when and under what conditions an interaction between two proteins or a transcription factor and its targets is likely to be realized. To advance solutions to this problem, more dynamic GRN data need to be collected and the AI methods that can track the dynamics of biological systems, like the ones of tracking objects and inferring actions in videos, need to be developed for GRN inference. AI agents that can conduct a series of reasoning and inference according to external inputs may also be applied to infer dynamic GRNs.